\documentclass{article}
\usepackage{ijcai26}

\usepackage{times}
\usepackage{soul}
\usepackage{url}
\usepackage[hidelinks]{hyperref}
\usepackage[utf8]{inputenc}
\usepackage[small]{caption}
\usepackage{graphicx}
\usepackage{amsmath}
\usepackage{amsthm}
\usepackage{booktabs}
\usepackage{algorithm}
\usepackage{algorithmic}
\usepackage{amssymb}      
\usepackage{amsfonts}
\usepackage{ijcai26}      
\usepackage{xcolor}

\usepackage[switch]{lineno}

\title{Constraint-Aware Aggregation for Federated Reinforcement Learning in Microgrid Energy Coordination}

\author{
Usman Haider$^1$
\and
Karl Mason$^{1,*}$\\
\affiliations
$^1$School of Computer Science, University of Galway, Ireland.\\
\emails
\{usman.haider, karl.mason\}@universityofgalway.ie
}

\begin{document}

\maketitle

\begin{abstract}
Federated Reinforcement Learning (FedRL) enables coordination of distributed energy resources without sharing raw local data, but standard aggregation methods such as FedAvg do not account for system-level constraints, often leading to unsafe global behavior. In this work, we study constraint-aware aggregation for federated reinforcement learning in distributed energy coordination. We propose aggregation rules that incorporate both local performance and estimated constraint violation into the server-side update. Among these, a simple penalty-based rule, $w_i \propto R_i - \alpha V_i$, consistently provides the most reliable trade-off between reward and safety, without requiring dual optimization or modifications to local training. \textcolor{black}{We evaluate our approach on DairyGridEnv, a benchmark modeling multiple farms coordinating battery storage under stochastic demand and a shared grid capacity constraint, and further assess robustness using real load-driven demand profiles from Finland and the German FIELD dataset. Across multiple seeds, penalty-based aggregation substantially reduces violations while improving reward relative to FedAvg in both synthetic and real load-driven settings.} A combined reward-violation scheme exposes a tunable trade-off via $\lambda$, but is less stable. These results demonstrate that lightweight aggregation strategies can substantially improve empirical safety in federated reinforcement learning while preserving standard communication protocols.
\end{abstract}

\section{Introduction}

Distributed Energy Resources (DERs), such as farm-level batteries and flexible electrical loads, are increasingly deployed in agricultural microgrids to reduce operating costs and improve energy resilience. In these systems, multiple farms interact through shared infrastructure, such as a common feeder or transformer, so local decisions can collectively violate global grid limits. This creates a constrained coordination problem in which each participant optimizes its own energy usage while remaining compatible with system-level safety requirements \cite{fedgrid2023,appliedenergy2025_frfl,engappai2024_frfl}.

Reinforcement Learning (RL) has shown strong potential for sequential decision-making in energy applications, including battery scheduling, demand response, and storage management under uncertainty \cite{egyai2025_frfl,appliedenergy2025_frfl}. However, most existing approaches rely on centralized training with access to globally aggregated trajectories. Such assumptions are often unrealistic in distributed agricultural settings, where farms may be unwilling or unable to share raw operational data due to privacy, ownership, or communication constraints \cite{engappai2024_frfl,appliedenergy2025_frfl}.

Federated Reinforcement Learning (FedRL) provides an appealing alternative by enabling decentralized training with periodic parameter aggregation, avoiding direct data sharing \cite{cheng2022_fedreview,grataloup2023_fedreview,arxiv2211_feddrl}. Recent work has applied FedRL to energy systems such as residential microgrids, electric vehicle charging, and distributed storage coordination, demonstrating benefits in scalability and privacy preservation \cite{appliedenergy2025_frfl,nature2025_fedev,energystorage_fedrl}. However, existing aggregation strategies largely inherit from FedAvg-style averaging, which treats all client updates uniformly and does not account for whether a policy induces constraint violations.

This limitation is critical in constrained energy systems. A client may achieve high local reward while contributing disproportionately to violations of shared infrastructure limits, making naive aggregation unsafe at the system level. Therefore, aggregation should depend not only on local performance but also on the compatibility of the induced behavior with global constraints. This motivates the need for constraint-aware aggregation mechanisms in federated reinforcement learning.

In this work, we study constraint-aware aggregation for FedRL. We use two lightweight client summaries, reward $R_i$ and violation estimate $V_i$, to reweight updates at the server without sharing trajectories or modifying local training. We consider several aggregation rules, including a simple penalty-based formulation $w_i \propto R_i - \alpha V_i$. We introduce DairyGridEnv, a constrained federated control benchmark where farms coordinate battery actions under stochastic demand, time-varying prices, and a shared grid limit. In addition to the synthetic setting, we evaluate using real farm load data from Finland \cite{uski2022data} and the FIELD dataset from Germany \cite{vavouris2025field}, thereby testing whether the observed trends generalize to realistic demand patterns.

Our results show that aggregation design strongly affects both reward and safety. In particular, penalty-based aggregation consistently achieves the best trade-off, significantly reducing violations while improving reward over FedAvg. A combined reward-violation scheme exposes a tunable trade-off but is less stable across seeds.

Our main contributions are as follows:
\begin{itemize}
\item We propose a constraint-aware aggregation framework for FedRL using reward and violation summaries without modifying local learning or communication.
\item We introduce DairyGridEnv and extend the evaluation to real-world farm-load datasets from Finland and Germany.
\item We empirically demonstrate that penalty-based aggregation provides a strong and stable reward-safety trade-off, substantially improving upon FedAvg and other aggregation strategies.
\item We show that these improvements can be achieved solely through server-side aggregation, highlighting aggregation design as a key lever for safety in federated reinforcement learning.
\end{itemize}
Overall, our results suggest that incorporating system-level constraints directly into aggregation is essential for deploying federated reinforcement learning in safety-critical energy systems.

\section{Related Work}

\textbf{Federated Reinforcement Learning.}
Federated Reinforcement Learning (FedRL) extends federated learning to sequential decision-making, allowing distributed agents to learn policies without sharing raw trajectories. Early methods commonly adopt FedAvg-style aggregation for policy or value-function parameters \cite{mcmahan2017communication,zhang2021federated_rl}, whereas recent work studies heterogeneity across agents and environments \cite{fedhql2023,fedrl_constraint_2024}. FedRL has also been applied to energy systems such as microgrids, electric vehicle charging, and distributed storage coordination \cite{appliedenergy2025_frfl,engappai2024_frfl,energystorage_fedrl}. However, most existing approaches do not explicitly account for shared system-level constraints during aggregation.

\textbf{Constraint-aware Federated Optimization.}
To address safety requirements, recent studies incorporate constrained optimization into federated learning using primal-dual or Lagrangian formulations. FedCM and related methods introduce dual variable exchange or constraint-gradient sharing to enforce feasibility \cite{fedcm2022}. While effective, these methods increase communication overhead and may expose sensitive constraint information, limiting their applicability in privacy-sensitive environments. More recent constrained federated reinforcement learning formulations consider distributed constrained MDPs and show that constraint heterogeneity significantly affects convergence and feasibility \cite{fedrl_constraint_2024}.

\textbf{Constrained Reinforcement Learning.}
Constrained reinforcement learning is commonly formulated as a constrained Markov decision process (CMDP), where policies maximize expected return subject to cost constraints \cite{achiam2017constrained}. Standard solutions use Lagrangian, primal-dual, or trust-region methods. In federated settings, constrained optimization methods such as FedCM introduce dual-variable exchange or constraint-gradient sharing to improve feasibility \cite{fedcm2022}. These methods can be effective, but they add communication and may expose constraint-related information. Recent constrained FedRL formulations further show that constraint heterogeneity affects convergence and feasibility \cite{fedrl_constraint_2024}.

\textbf{Energy Systems and Multi-Agent Coordination.}
RL has been widely used for battery scheduling, demand response, and microgrid control \cite{waghmare2025systematic}. Existing dairy-farm battery management studies mainly focus on single-farm control using conventional RL or rule-based strategies, often with real load and renewable profiles. In contrast, our work studies multi-farm coordination under a shared feeder constraint. Multi-agent RL can address such coordination problems, but many approaches require centralized critics, shared experience, or full observability.

\textbf{Contribution of Our Work.}
Unlike prior FedRL methods that rely on uniform aggregation or explicit dual updates, we study constraint-aware aggregation as a simple server-side mechanism for balancing performance and safety. Our approach leaves the local RL objective and local policy optimization procedure unchanged and uses only lightweight reward and violation summaries for aggregation. It therefore adds only negligible communication cost beyond the standard federated model exchange while remaining compatible with standard policy optimization methods, assuming coordinator-level computation of the global violation signal. This makes it attractive for privacy-sensitive deployments that require scalable and robust coordination.

\section{Problem Setting: DairyGridEnv}

We introduce DairyGridEnv, a constrained multi-agent reinforcement learning environment modeling coordination among dairy farms sharing a rural grid connection. The environment captures the central challenge addressed in this work: each agent optimizes local battery operation under partial observations, while all agents are coupled through a shared grid capacity constraint.
We consider $N=5$ farms connected to a common feeder with a capacity limit $L=12$. Here, $N$ denotes the number of farms (agents), while $L$ represents the feeder capacity in normalized power units. All power-related quantities in the environment (demand, actions, and grid load) are expressed in a common normalized unit rather than physical units such as kW to focus on coordination behavior.

\subsection{Local Observations and Actions}

At each timestep $t$, agent $i$ observes
\[
o_i^t = (p_t, s_i^t, d_i^t),
\]
where $p_t = 10 + 4\sin(t/8)$ denotes a normalized time-varying electricity cost, $s_i^t \in [0,1]$ is the battery state-of-charge (SOC), and $d_i^t$ is the electricity demand of farm $i$. In the controlled synthetic benchmark, demand is generated as
\[
d_i^t = d_{\mathrm{base}} + 0.6\sin(t/5) + \eta_i, \quad \eta_i \sim U(-0.5,0.5),
\]
where the sinusoidal component models periodic consumption and $\eta_i$ captures demand fluctuations. In all synthetic experiments, we set $d_{\mathrm{base}} = 1.5$.

Each agent selects a continuous action.
\[
a_i^t \in [-2,2],
\]

where positive values correspond to battery charging and negative values correspond to discharging. The bounds $[-2,2]$ define the maximum allowable charge and discharge rates relative to the normalized demand scale. The state of the battery is updated as:
\[
s_i^{t+1} = \mathrm{clip}(s_i^t + 0.1 a_i^t,\, 0,\, 1),
\]
which enforces valid SOC bounds.

\subsection{Synthetic and Real Load Profiles}
\textcolor{black}{
The main controlled benchmark uses the synthetic demand process above to isolate the effect of aggregation under a known stochastic environment. To evaluate robustness under real demand variability, we additionally use two real load-driven settings. First, the Finland profile contains farm load, photovoltaic generation, and wind generation. We construct net demand as
\[
d_{\mathrm{net}}^t = \max\left(\mathrm{Load}^t - \mathrm{PV}^t - \mathrm{Wind}^t, 0\right),
\]
and normalize it to the simulator demand range.
Second, we use the German FIELD dataset, which contains one-second three-phase electrical load measurements from 30 dairy farms in Germany over more than one year \cite{vavouris2025field}. We compute farm demand as the sum of the three phases,
\[
d_i^t = \mathrm{phase1}_i^t + \mathrm{phase2}_i^t + \mathrm{phase3}_i^t,
\]
resample the data to hourly resolution, select five farms with diverse average load levels, and normalize each demand profile to match the operating range of DairyGridEnv.
The real load datasets do not include electricity price data. Therefore, all real load-driven experiments retain the same normalized time-varying price $p_t$ used in the synthetic benchmark. Thus, real data are used to drive demand variability, while price remains controlled.}

\subsection{Grid Coupling and Constraint}
Agents are coupled through a shared grid constraint. The local net grid contribution of agent $i$ at time $t$ is
\[
g_i^t = d_i^t + \max(a_i^t,0) - \max(-a_i^t,0),
\]
which represents the net power drawn from the grid by agent $i$. The term $d_i^t$ denotes local electricity demand; charging increases grid consumption, while discharging reduces it by supplying energy from the battery.
The total grid import is then:
\[
g_t = \sum_{i=1}^{N} g_i^t
= \sum_{i=1}^{N} \left(d_i^t + \max(a_i^t,0) - \max(-a_i^t,0)\right).
\]

A violation occurs when the feeder capacity is exceeded:
\[
\mathrm{viol}^t = \max(0, g_t - L).
\]

This imposes a global constraint under decentralized decision-making, which constitutes the main challenge in the federated setting.

\subsection{Violation Attribution for Aggregation}

Assigning responsibility for global violations to individual agents is non-trivial under decentralized observations. In the implementation, we use a heuristic attribution rule based on each client's net grid contribution:
\[
V_i^t = \mathrm{viol}^t \cdot \frac{g_i^t}{\sum_{j=1}^{N} g_j^t + \epsilon},
\qquad \epsilon = 10^{-6}.
\]

Equivalently, this can be written as
\[
V_i^t
=
\mathrm{viol}^t
\cdot
\frac{d_i^t + \max(a_i^t,0) - \max(-a_i^t,0)}
{\sum_{j=1}^{N} \left(d_j^t + \max(a_j^t,0) - \max(-a_j^t,0)\right) + \epsilon}.
\]

This attribution uses directional net-grid contributions rather than absolute action magnitudes, so discharging actions that reduce net import receive a smaller attributed violation signal. However, the rule is still heuristic and should be interpreted as a lightweight proxy for aggregation rather than an exact physical or causal decomposition of responsibility. In particular, clients with net export may receive negative attributed shares under this formulation.

In our framework, $V_i^t$ is computed at the environment or coordinator level after the joint action is observed and is used solely as a scalar summary for server-side aggregation.

\subsection{Reward Design}

Each agent receives a local reward consisting of its own energy cost contribution together with a shared penalty for violating the feeder limit:
\[
r_i^t = -p_t \, g_i^t - \frac{100}{N}\,\mathrm{viol}^t.
\]

Equivalently,
\[
r_i^t
=
-p_t \left(d_i^t + \max(a_i^t,0) - \max(-a_i^t,0)\right)
-
\frac{100}{N}\,\mathrm{viol}^t.
\]

Thus, all agents share the same violation penalty term but differ in the local energy cost term, which depends on their net grid contribution. Episodes last $T=40$ steps.

\subsection{Design Rationale}

DairyGridEnv is designed as a compact benchmark for constrained federated control. It combines three key elements:
(i) decentralized observations,
(ii) a shared global constraint,
and (iii) stochastic demand dynamics.

This structure isolates the effect of aggregation on the reward-safety trade-off without introducing unnecessary system complexity.

\begin{figure*}[t]
\centering
\includegraphics[width=0.95\textwidth]{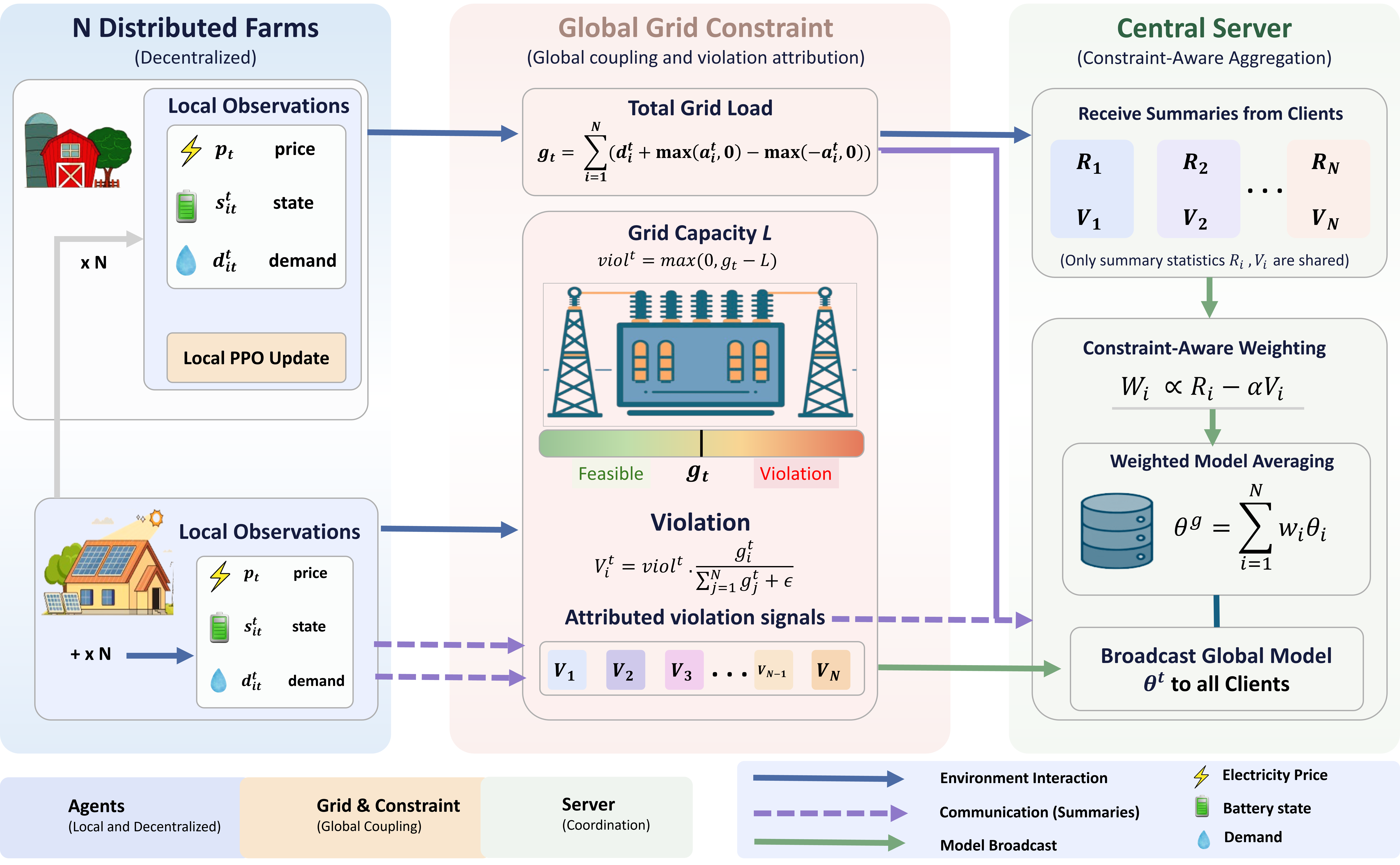}
\caption{\textbf{Constraint-aware federated reinforcement learning framework.} Each farm trains a local PPO policy from private observations $(p_t, s_i^t, d_i^t)$, while all farms are coupled through a shared grid constraint. Joint actions determine the total grid load $g_t$ and the global violation $\mathrm{viol}^t=\max(0,g_t-L)$. After each rollout, the central server uses the heuristic-attribute violation summaries $V_i$ and average rewards $R_i$ to compute constraint-aware aggregation weights and to form the global model $\theta^g$, which is then broadcast back to all clients.}
\label{fig:framework}
\end{figure*}

\section{Constraint-Aware Federated Reinforcement Learning}

\subsection{System Overview}

Figure~\ref{fig:framework} illustrates the proposed framework. The system consists of distributed clients and a central server that interact through a shared, constrained environment. Each client observes only local state $o_i^t = (p_t, s_i^t, d_i^t)$ and selects an action $a_i^t \in [-2,2]$.

Clients do not share trajectories. Instead, after each rollout, they report two scalar summaries:
\[
R_i = \frac{1}{T}\sum_{t=1}^{T} r_i^t,
\qquad
V_i = \frac{1}{T}\sum_{t=1}^{T} V_i^t.
\]

The server aggregates local models according to
\[
\theta^g = \sum_i w_i \theta_i,
\qquad
\sum_i w_i = 1,
\]
where the weights depend on $(R_i, V_i)$.

A key challenge is that the shared grid constraint depends on the joint action and is not directly observable locally. In the current implementation, the global violation and the attributed violation summaries are computed at the environment or coordinator level after observing the joint action. Thus, the method avoids sharing raw trajectories or local state histories, but it does assume synchronized rollout execution and access to joint actions for computing the global safety signal. The attribution mechanism, therefore, provides a heuristic proxy signal that enables constraint-aware aggregation without sharing full trajectories.

\subsection{Federated Problem Setup}

Each client interacts with the shared environment, where joint actions determine the local net grid contributions.
\[
g_i^t = d_i^t + \max(a_i^t,0) - \max(-a_i^t,0),
\]
the total grid load
\[
g_t = \sum_{i=1}^{N} g_i^t,
\]
and the resulting violation
\[
\mathrm{viol}^t = \max(0, g_t - L).
\]

After each rollout, clients transmit $(R_i, V_i)$ together with updated parameters $\theta_i$ to the server. Here,
\[
R_i = \frac{1}{T}\sum_{t=1}^{T} r_i^t,
\qquad
V_i = \frac{1}{T}\sum_{t=1}^{T} V_i^t.
\]

Although each client acts only on local observations, rollout generation occurs in a shared environment in which all client actions jointly determine the global constraint signal.

\subsection{Local Policy Optimization}

Each client parameterizes its policy as
\[
\pi_{\theta_i}(o_i^t) = 2 \cdot \tanh(\mathrm{MLP}_{3\to64\to64\to1}(o_i^t)).
\]
Policies are trained locally using a clipped PPO update with a discount factor $\gamma=0.99$ and a learning rate of $10^{-4}$. Advantages are computed from discounted returns and normalized within each local rollout. At each round, clients initialize from $\theta^g$, collect a rollout of length $T=40$, and perform local updates.

\subsection{Constraint-Aware Aggregation}

We consider aggregation rules that depend on reward and violation summaries. 
The global model is computed as
\[
\theta^g = \sum_{i=1}^{N} w_i \theta_i,
\qquad
\sum_{i=1}^{N} w_i = 1.
\]
Weights for each rule are shown in Table \ref{tab:agg}.

\begin{table}[H]
\centering
\caption{Constraint-aware aggregation rules. All weights are normalized across clients before model averaging.}
\label{tab:agg}
\begin{small}
\begin{tabular}{ll}
\toprule
Method & Weighting Rule \\
\midrule
Performance & $w_i \propto \widetilde{R}_i$ \\
Violation & $w_i \propto \frac{1}{1+V_i}$ \\
Combined & $w_i \propto \frac{\widetilde{R}_i+\epsilon}{1+\lambda \widetilde{V}_i}$ \\
Penalty & $w_i \propto \max\!\big(\epsilon,\; R_i-\alpha V_i-\min_j(R_j-\alpha V_j)\big)$ \\
\bottomrule
\end{tabular}
\end{small}
\end{table}
The penalty formulation directly trades off performance and constraint violation without introducing additional optimization variables. In all reported experiments, we fix $\alpha=1.0$. For the combined rule, $\widetilde{R}_i$ and $\widetilde{V}_i$ denote min-max normalized reward and violation summaries within each communication round. The small constant $\epsilon$ prevents zero weights and division by zero.




\section{Experiments}
We first evaluate on the synthetic DairyGridEnv benchmark with $N=5$ farms, grid limit $L=12.0$, horizon $T=40$, and demand base set to $1.5$. All quantities are expressed in normalized units, with demand, actions, and grid capacity sharing a common scale. Each local controller uses an MLP policy, $\mathrm{MLP}(3 \to 64 \to 64 \to 1,\tanh)$, trained with PPO using clipping parameter $\epsilon=0.2$, discount factor $\gamma=0.99$, and learning rate $10^{-4}$. Federated training runs for 30 communication rounds. For each method, we evaluate over 10 episodes per seed across 5 seeds $\{0,\dots,4\}$ and report mean and standard deviation across seeds. We additionally report centralized PPO baselines under the same demand setting. For penalty-based aggregation, we fix $\alpha=1.0$ in all reported experiments.

We report four evaluation metrics. Rewards represent negative operating cost, so larger values, i.e., values closer to zero, indicate better performance. Reward denotes the mean cumulative per-step reward across clients over an evaluation episode. Mean violation denotes the mean across clients of cumulative attributed violation over an evaluation episode. MaxViol denotes the maximum single-step global violation within an evaluation episode. The violation rate denotes the fraction of clients with a nonzero cumulative attributed violation in an evaluation episode. Violation and MaxViol are reported in normalized power units, while Reward reflects normalized operating cost.

\subsection{Hyperparameter Ablation: \texorpdfstring{$\lambda$}{lambda} Selection for Combined Aggregation}
We first analyze the trade-off parameter $\lambda$ in the combined aggregation rule. We sweep $\lambda \in \{0.5, 0.75, \dots, 3.0\}$ and measure the resulting reward and constraint violation. The sweep reveals a clear trade-off between reward and safety. Small values of $\lambda$ underweight violations and lead to poor safety, for example, $\lambda=0.5$ yields a mean violation of $10.19 \pm 5.09$. Increasing $\lambda$ reduces violations substantially. The best safety-oriented operating point is at $\lambda=1.5$, which achieves the lowest mean violation, $0.90 \pm 1.74$, while also maintaining a competitive reward, $-15.99 \pm 10.68$. Nearby values such as $\lambda=2.0$ and $\lambda=2.25$ are also competitive, but $\lambda=1.5$ provides the most favorable balance under our selection criterion. In Figure 2, reward is plotted in absolute magnitude for visual clarity, while Table 2 reports the original signed reward values.
\begin{figure}[!htbp]
\centering
\includegraphics[width=0.52\textwidth]{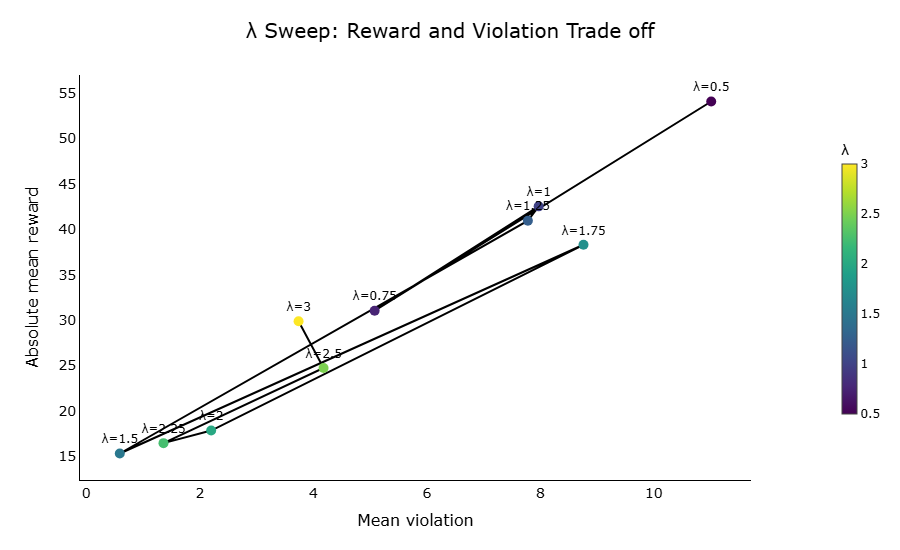}
\caption{ $\lambda$ sweep for the combined aggregation rule. Each point corresponds to a different value of $\lambda$. The horizontal axis shows mean violation, and the vertical axis shows absolute mean reward magnitude. Moderate values yield a better balance between reward and safety than either very small or very large values.}
\label{fig:lambda}
\end{figure}
\begin{figure*}[t]
\centering
\includegraphics[width=\textwidth]{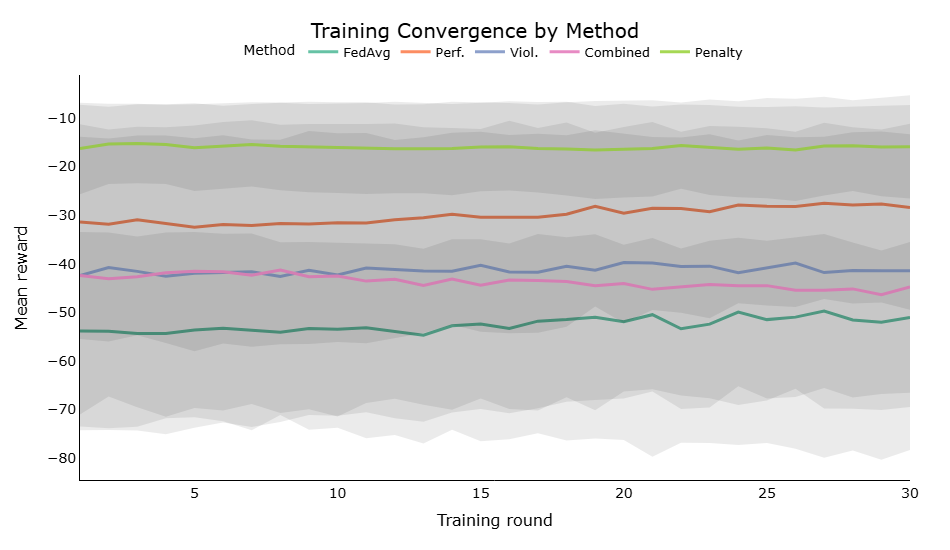}
\caption{Training convergence across communication rounds. Penalty-based and performance-weighted aggregation achieve consistently better reward trajectories than FedAvg and violation-weighted aggregation.}
\label{fig:convergence}
\end{figure*}

These results show that the combined rule exposes a tunable reward-safety frontier, but its effectiveness depends strongly on the choice of $\lambda$.

\subsection{Federated Learning Results}
We compare five aggregation strategies: Federated Averaging (FedAvg), Performance-weighted aggregation (Perf), Violation-weighted aggregation (Viol), Combined reward-and-violation-weighted aggregation (Combined) with tuned $\lambda$, and Penalty-based aggregation (Penalty).
\begin{table}[t]
\centering
\caption{Effect of $\lambda$ on the reward and violation trade-off for the combined rule.}
\label{tab:lambda}
\scriptsize
\resizebox{0.9\columnwidth}{!}{%
\begin{tabular}{lcc}
\toprule
$\lambda$ & Reward & Violation \\
\midrule
0.5  & $-52.01 \pm 15.77$ & $10.19 \pm 5.09$ \\
0.75 & $-28.44 \pm 21.11$ & $4.30 \pm 5.35$ \\
1.0  & $-42.01 \pm 27.91$ & $7.68 \pm 8.61$ \\
1.25 & $-44.79 \pm 33.62$ & $9.21 \pm 9.81$ \\
1.5  & $\mathbf{-15.99 \pm 10.68}$ & $\mathbf{0.90 \pm 1.74}$ \\
1.75 & $-37.31 \pm 36.24$ & $8.49 \pm 8.96$ \\
2.0  & $-15.55 \pm 17.51$ & $1.76 \pm 3.52$ \\
2.25 & $-14.79 \pm 13.51$ & $1.16 \pm 2.31$ \\
2.5  & $-23.71 \pm 28.00$ & $3.97 \pm 7.52$ \\
3.0  & $-26.34 \pm 17.17$ & $2.83 \pm 4.88$ \\
\bottomrule
\end{tabular}%
}
\end{table}

Figure~\ref{fig:convergence} shows that aggregation choice strongly affects convergence behavior.  Penalty-based aggregation maintains the strongest average reward profile across rounds and exhibits more reliable behavior than the baselines. Performance weighted aggregation is the next strongest method. In contrast, FedAvg and violation weighted aggregation remain substantially worse throughout training. The combined rule improves over the weakest baselines in some rounds, but exhibits higher variability.

\begin{figure}[t]
\centering
\includegraphics[width=0.5\textwidth]{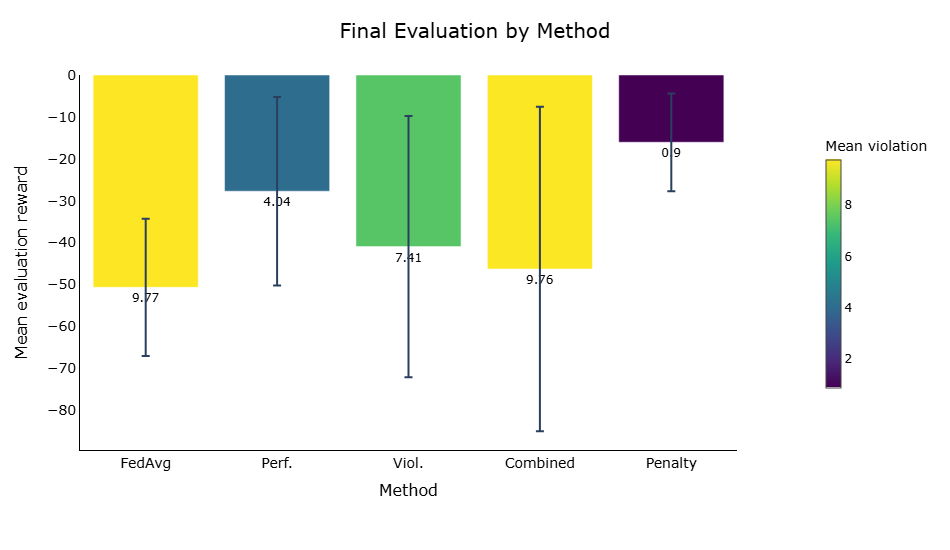}
\caption{Federated evaluation on the synthetic benchmark. Bar height shows the mean reward, and color indicates the mean violation.}
\label{fig:eval}
\end{figure}

Figure~\ref{fig:eval} confirms the same pattern at evaluation time. Among federated methods, penalty-based aggregation achieves the best overall trade-off, with the highest reward and the lowest violation. Performance-weighted aggregation is the second-strongest method. FedAvg and violation-weighted aggregation perform poorly, with high violations across seeds. Although the combined rule is promising in the $\lambda$ sweep, its multi-seed performance is less stable than penalty-based aggregation.

Table~\ref{tab:main} shows that penalty-based aggregation is the strongest federated method. Relative to FedAvg, it reduces mean violation from $9.77$ to $0.90$, while also improving mean reward from $-50.71$ to $-16.06$. Performance-weighted aggregation is a competitive baseline but remains clearly behind penalty-based aggregation in both reward and safety. The combined rule, despite having a favorable operating point in the $\lambda$ sweep, is not as robust across seeds in the final comparison.

\begin{table}[!htbp]
\centering
\caption{Federated evaluation across 5 seeds. Values are reported as mean $\pm$ sample standard deviation across seeds.}
\label{tab:main}
\scriptsize
\setlength{\tabcolsep}{4pt}
\resizebox{\columnwidth}{!}{%
\begin{tabular}{lcccc}
\toprule
Method & Reward & Violation & MaxViol & Voilation Rate \\
\midrule
FedAvg   & $-50.71 \pm 16.36$ & $9.77 \pm 5.21$ & $4.57 \pm 1.50$ & $1.00 \pm 0.00$ \\
Perf     & $-27.75 \pm 22.53$ & $4.04 \pm 5.56$ & $1.92 \pm 2.64$ & $0.40 \pm 0.55$ \\
Viol     & $-40.98 \pm 31.17$ & $7.41 \pm 9.62$ & $3.04 \pm 3.05$ & $0.80 \pm 0.45$ \\
Combined & $-46.34 \pm 38.73$ & $9.76 \pm 11.32$ & $3.81 \pm 3.02$ & $0.80 \pm 0.45$ \\
\textbf{Penalty} & $\mathbf{-16.06 \pm 11.67}$ & $\mathbf{0.90 \pm 1.79}$ & $\mathbf{0.79 \pm 1.33}$ & $\mathbf{0.40 \pm 0.55}$ \\
\bottomrule
\end{tabular}%
}
\end{table}

\subsection{Centralized Reference Baselines}
We also report centralized PPO baselines under the same demand setting, namely centralized PPO, PPO with a Lagrangian update, and PPO with a fixed penalty. Table~\ref{tab:best_vs_baseline} compares our best federated method, penalty-based aggregation, with these centralized baselines under the same environment setting. Penalty aggregation achieves the strongest reward-violation trade-off among federated methods, substantially reducing constraint violations compared to standard federated aggregation.

\begin{table}[!htbp]
\centering
\caption{Best federated method vs. centralized baselines. Penalty aggregation is the strongest federated method, while centralized PPO provides a reference upper bound under full coordination.}
\label{tab:best_vs_baseline}
\scriptsize
\setlength{\tabcolsep}{4pt}
\resizebox{\columnwidth}{!}{%
\begin{tabular}{lcccc}
\toprule
Method & Reward & Violation & MaxViol & Voilation Rate \\
\midrule
\textbf{Ours (Penalty)} & $\mathbf{-16.06 \pm 11.67}$ & $\mathbf{0.90 \pm 1.79}$ & $\mathbf{0.79 \pm 1.33}$ & $\mathbf{0.40 \pm 0.55}$ \\
PPO         & $-97.85 \pm 2.014$ & $0.0223 \pm 0.0145$ & $0.41 \pm 0.23$ & N/A \\
PPO-Lag     & $-96.85 \pm 2.104$ & $0.0253 \pm 0.0168$ & $0.46 \pm 0.27$ & N/A \\
PPO-Penalty & $-98.84 \pm 2.112$ & $0.0226 \pm 0.0135$ & $0.43 \pm 0.21$ & N/A \\
\bottomrule
\end{tabular}%
}
\end{table}
The centralized baselines achieve near-zero violations due to full access to the global system state and joint action optimization. This highlights that the primary challenge in the federated setting is not the control task itself, but enforcing constraints under decentralized updates and server-side aggregation. The centralized results also indicate that this benchmark is relatively easy under full coordination, since centralized PPO, PPO with a Lagrangian update, and PPO with a fixed penalty converge to nearly identical performance. Absolute reward values are not directly comparable across centralized and federated settings because the centralized controller operates on the joint system state. In contrast, the two regimes use different control formulations and reporting scales. We therefore compare centralized and federated settings primarily through violations and qualitative control difficulty, rather than raw reward magnitude.
\subsection{Evaluation with Real Load Profiles}
\textcolor{black}{To assess whether the observed trends persist beyond the synthetic demand process, we additionally evaluate the aggregation rules using real farm load profiles. For the Finland profile, demand is constructed from farm load, PV generation, and wind generation. For the German FIELD dataset, each aggregate CSV file corresponds to a single farm and contains three-phase load measurements \cite{vavouris2025field}. We compute total farm load as the sum of the three phases, resample the measurements to hourly resolution, select five farms, and normalize each profile to the demand range used in DairyGridEnv. 
\begin{table}[t]
\centering
\scriptsize
\setlength{\tabcolsep}{2.5pt} 
\caption{Evaluation with real load profile driven demand. Values are mean $\pm$ std across five seeds.}
\label{tab:real_loads}
\begin{tabular}{llcccc}
\toprule
Data & Method & Reward & Violation & MaxViol & Violation Rate \\
\midrule
Finland & FedAvg & $-59.08 \pm 23.25$ & $12.90 \pm 7.50$ & $5.48 \pm 2.03$ & $0.98 \pm 0.04$ \\
Finland & Perf & $-38.71 \pm 36.73$ & $7.75 \pm 10.73$ & $2.95 \pm 3.76$ & $0.58 \pm 0.40$ \\
Finland & Viol & $-43.41 \pm 37.67$ & $8.54 \pm 11.96$ & $3.60 \pm 3.55$ & $0.78 \pm 0.39$ \\
Finland & Comb & $-52.92 \pm 40.27$ & $12.13 \pm 11.59$ & $4.51 \pm 3.29$ & $0.80 \pm 0.45$ \\
Finland & \textbf{Pen} & $\mathbf{-20.08 \pm 15.62}$ & $\mathbf{2.26 \pm 2.95}$ & $\mathbf{1.55 \pm 1.82}$ & $\mathbf{0.52 \pm 0.43}$ \\
\midrule
FIELD & FedAvg & $-23.02 \pm 5.88$ & $0.92 \pm 0.83$ & $0.69 \pm 0.52$ & $0.60 \pm 0.38$ \\
FIELD & Perf & $-14.34 \pm 12.79$ & $0.79 \pm 1.77$ & $0.43 \pm 0.95$ & $0.22 \pm 0.44$ \\
FIELD & Viol & $-20.39 \pm 14.80$ & $1.41 \pm 3.07$ & $0.61 \pm 1.21$ & $0.36 \pm 0.46$ \\
FIELD & Comb & $-20.34 \pm 17.54$ & $1.61 \pm 3.03$ & $0.81 \pm 1.38$ & $0.40 \pm 0.55$ \\
FIELD & \textbf{Pen} & $\mathbf{-8.27 \pm 8.45}$ & $\mathbf{0.00 \pm 0.01}$ & $\mathbf{0.01 \pm 0.02}$ & $\mathbf{0.04 \pm 0.05}$ \\
\bottomrule
\end{tabular}
\end{table}
Table~\ref{tab:real_loads} shows that the main conclusion remains consistent under real load-driven demand. Penalty-based aggregation achieves the best overall reward-safety trade-off in both real data settings. On the Finland profile, penalty reduces mean violation from $12.90$ under FedAvg to $2.26$. On the German FIELD profile, penalty nearly eliminates violations, reducing the mean violation from $0.92$ under FedAvg to $0.00$. These results suggest that the benefit of constraint-aware aggregation is not limited to the synthetic demand process.}

\subsection{Statistical Comparison}
For the synthetic benchmark, all methods are evaluated on the same five seeds. We perform paired statistical tests for the strongest comparison, namely Penalty versus FedAvg. For reward, penalty significantly improves over FedAvg under both a one-sided paired $t$-test ($t=3.48$, $p=0.0126$) and a Wilcoxon signed-rank test ($p=0.0313$). For mean violation, penalty significantly reduces constraint violations relative to FedAvg ($t=-3.71$, $p=0.0103$, Wilcoxon $p=0.0313$). Similarly, for maximum violation, penalty achieves significantly lower values ($t=-4.73$, $p=0.0045$, Wilcoxon $p=0.0313$). Given the small number of seeds, we restrict statistical claims to this primary comparison and avoid over-interpreting pairwise differences between all methods.




\section{Conclusion}

We studied constraint-aware aggregation for federated reinforcement learning in distributed energy coordination. Using DairyGridEnv, we showed that aggregation design strongly affects both reward and constraint satisfaction. We further evaluated the methods under real load-driven demand profiles from Finland and the German FIELD dataset. Across these settings, penalty-based aggregation offers the most reliable average reward-safety trade-off among the methods considered, substantially improving over FedAvg in both reward and violation behavior. While combined reward-violation weighting offers a tunable trade-off via $\lambda$, it is less stable across seeds. A comparison with centralized baselines shows that enforcing constraints in decentralized learning and aggregation remains a central challenge. Overall, our results suggest that simple server-side aggregation changes can substantially improve empirical safety without modifying local training. Future work will investigate larger heterogeneous systems, real price signals, asynchronous participation, and stronger constrained federated RL baselines.


\bibliographystyle{named}
\bibliography{ijcai26}

\end{document}